\documentclass[final,3p,times,twocolumn,authoryear]{elsarticle}

\usepackage{amssymb}
\usepackage{amsmath}

\usepackage{subcaption}
\usepackage{hyperref}
\usepackage{url}
\hypersetup{breaklinks=false}

\journal{Elesevier}

\begin{document}

\begin{frontmatter}



\title{Learning Street View Representations with Spatiotemporal Contrast
} 


\author[label1,label2]{Yong Li\corref{equalcontrib}} 

\author[label1]{Yingjing Huang\corref{equalcontrib}}

\author[label3]{Gengchen Mai}

\author[label2]{Fan Zhang\corref{cor1}}
\ead{fanzhanggis@pku.edu.cn}
\cortext[equalcontrib]{ Equal contribution}
\cortext[cor1]{Corresponding author}

\affiliation[label1]{organization={Institute of Remote Sensing and Geographic Information System, School of Earth and Space Sciences, Peking University},
	                city={Beijing},
                     postcode={100871}, 
                     country={China}}

\affiliation[label2]{organization={Department of Civil and Environmental Engineering, The Hong Kong University of Science and Technology, Hong Kong SAR, China}}
\affiliation[label3]{organization={SEAI Lab, Department of Geography and the Environment, University of Texas at Austin},
            city={Austin},
            postcode={78712},
            state={Texas},
            country={USA}}

\begin{abstract}
Street view imagery is extensively utilized in representation learning for urban visual environments, supporting various sustainable development tasks such as environmental perception and socio-economic assessment. 
However, it is challenging for existing image representations to specifically encode the dynamic urban environment (such as pedestrians, vehicles, and vegetation), the built environment (including buildings, roads, and urban infrastructure), and the environmental ambiance (such as the cultural and socioeconomic atmosphere) depicted in street view imagery to address downstream tasks related to the city.
In this work, we propose an innovative self-supervised learning framework that leverages temporal and spatial attributes of street view imagery to
learn image representations of the dynamic urban environment 
for diverse downstream tasks. By employing street view images captured at the same location over time and spatially nearby views at the same time, we construct contrastive learning tasks designed to learn the temporal-invariant characteristics of the built environment and the spatial-invariant neighborhood ambiance. Our approach significantly outperforms traditional supervised and unsupervised methods in tasks such as visual place recognition, socioeconomic estimation, and human-environment perception. 
Moreover, we demonstrate the varying behaviors of image representations learned through different contrastive learning objectives across various downstream tasks. This study systematically discusses representation learning strategies for urban studies based on street view images, providing a benchmark that enhances the applicability of visual data in urban science. 
The code is available at \href{https://github.com/yonglleee/UrbanSTCL}{https://github.com/yonglleee/UrbanSTCL}.
\end{abstract}




\end{frontmatter}




\section{Introduction}

In recent years, unsupervised learning has demonstrated outstanding performance on various downstream tasks. By leveraging methods such as contrastive learning~\citep{he2020Momentum,chen2020Simple,chen2021empirical} and masked image modeling~\citep{he2022masked,xie2022simmim}, it has achieved efficient image representation and exhibited excellence in classical computer vision tasks such as image classification~\citep{radford2021learning}, object detection~\citep{he2022masked}, and semantic segmentation~\citep{wang2020self}, surpassing the vast majority of traditional supervised learning only approaches. However, current unsupervised learning and self-supervised learning aim to encode as much semantic and structural information of objects and environments in a scene as possible~\citep{park2023what,huang2024how}. This is not suitable for all downstream tasks in tasks such as street view-based urban environment understanding. For instance, in place recognition tasks~\citep{lowry2015visual}, the features are expected to focus only on place-invariant information, such as buildings and roads, filtering out dynamic information like lighting conditions, pedestrians, vehicles, and vegetation. In contrast, in tasks related to human perception of places~\citep{dubey2016deep,zhang2018measuring}, these dynamic elements are important. Moreover, tasks like socioeconomic prediction~\citep{wang2020urban2vec} emphasize the spatially consistent expression of neighboring scenes.

In image representation learning, selectively encoding dynamic and static information in urban environments and the ambiance they create is highly important but inherently challenging~\citep{cordts2016cityscapes}. Achieving precise encoding of such information typically requires separately labeling dynamic and static elements and using specific training strategies (e.g., masking dynamic elements when encoding static ones~\citep{cheng2017segflow,wang2019learning}). However, both the labeling and training processes are fraught with difficulties. Factors such as lighting conditions, vegetation appearance, and ground litter are challenging to label objectively and consistently. This makes it nearly impossible to accurately represent these complex environmental factors using traditional datasets (e.g., ImageNet~\citep{deng2009imagenet}, Places~\citep{zhou2017places}) and classical methods (supervised or unsupervised).

\begin{figure*}[ht!]
\begin{center}
\includegraphics[width=\linewidth]{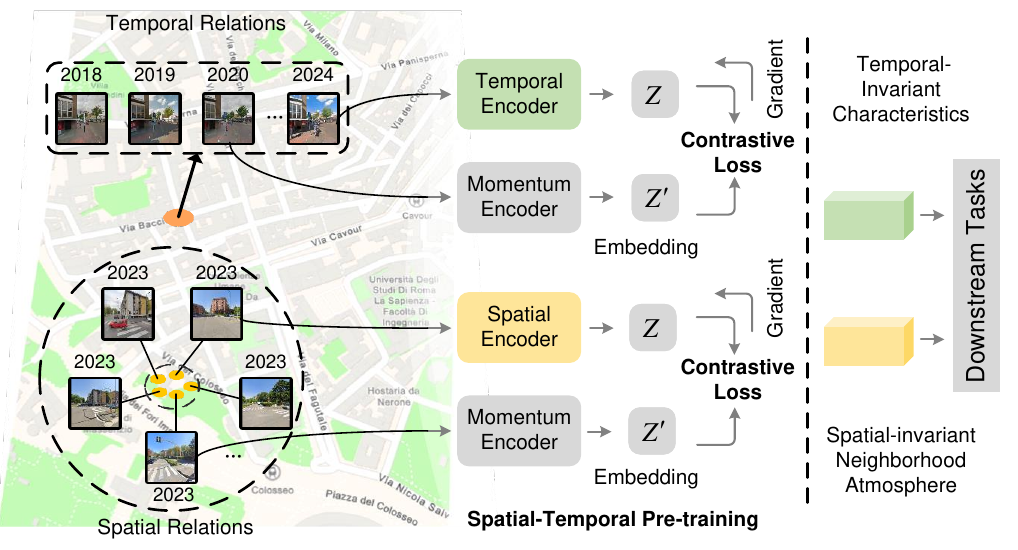}
\end{center}
\caption{Spatial and temporal contrastive learning with street view images. Using street view images captured at the same location over time, contrastive learning tasks are designed to learn the temporal-invariant characteristics of the built environment; Using spatially proximate street view images from the same period, learning tasks are crafted to learn the spatial-invariant neighborhood ambiance, such as socioeconomic atmosphere. }
\label{fig:framework}
\end{figure*}

Unlike existing large-scale datasets, street view imagery, as a high-resolution urban visual dataset, possesses unique spatiotemporal attributes that can capture both dynamic and static information in urban environments and the ambiance they form~\citep{biljecki2021street, zhang2024urban}. Therefore, in this work, we leverage these spatiotemporal attributes of street view imagery to propose a self-supervised urban visual representation framework based on street view imagery (see Figure~\ref{fig:framework}). This framework aims to selectively extract and encode dynamic and static elements and their ambiance in urban environments according to the requirements of different downstream tasks, achieving precise representations of urban environments. Specifically, the framework is based on the following three hypotheses:

\begin{itemize}
    \item \textbf{Temporal Invariance Representation}: At the same location, static elements such as buildings and streets do not change in images taken at different times, whereas dynamic elements like lighting conditions, pedestrians, vehicles, and vegetation present randomness in images taken at different times. Learning temporal invariant representations can retain the encoding of static elements while automatically filtering out information about dynamic elements. To capture this temporal invariance, we utilize the temporal attributes of street view imagery to construct positive sample pairs from historical street view images taken at different times at the same location. We expect that, after pre-training, the temporal encoder can learn stable features of the built environment. This method is suitable for tasks that rely on temporal stability, such as visual place recognition.
    \item \textbf{Spatial Invariance Representation}: At the same time, urban spaces at nearby locations usually exhibit similarity; the architectural styles and urban functions in adjacent areas are relatively consistent, while specific visual elements in images of nearby locations present randomness. Learning spatial invariant representations can encode the overall neighborhood ambiance within a specific spatial range while avoiding focusing on any specific elements. To capture this spatial invariance, we leverage the spatial attributes of street view imagery to construct positive sample pairs from street view images taken in adjacent areas at the same time. We expect that, after pre-training, the spatial encoder can learn spatially invariant neighborhood ambiance. This method is suitable for tasks that require spatial consistency, such as socioeconomic prediction.
    \item \textbf{Global Information Representation}: Besides temporal and spatial invariance, there are elements in urban environments that require holistic perception; these global features are vital for tasks involving human perception. To capture these characteristics, we construct positive sample pairs by applying data augmentation to the same street view image. We expect that, after pre-training, the model can retain the key elements of the scene and comprehensively capture the global information of the image.
\end{itemize}

We validate the effectiveness of these hypotheses across multiple urban downstream tasks. Experimental results demonstrate that different contrastive learning objectives can learn different types of features that are more suitable for their respective downstream tasks. We also conduct an in-depth analysis of the reasons behind the performance of different contrastive methods, further underscoring the importance of targeted learning strategies. This study systematically explores representation learning strategies in urban studies based on street view images, provides a valuable benchmark, and enhances the applicability of visual data in urban science.

\section{Related work}

\subsection{Self-Supervised Representation Learning}
Self-supervised representation learning leverages the inherent structure within data to generate supervisory signals, thereby mitigating the need for extensive labeled datasets. A prominent approach in this field is contrastive learning, which has demonstrated significant success in learning robust representations. Methods such as InstDis \citep{wu2018unsupervised}, SimCLR \citep{chen2020Simple}, and the MoCo series \citep{he2020Momentum, chen2021empirical} focus on contrasting positive pairs of similar instances against negative pairs of dissimilar instances to learn effective features. In contrast, BYOL \citep{grill2020bootstrap}, SimSiam \citep{chen2021Exploring}, and DINO \citep{caron2021emerging} improve performance by avoiding negative samples altogether and adopting a self-distillation approach. These methods have achieved notable results in various visual tasks, such as image classification and object detection, showcasing the effectiveness of self-supervised learning to perform exceptionally well with large-scale unlabeled data. However, despite these successes, existing self-supervised learning methods predominantly focus on static images without considering the spatiotemporal context inherent in certain datasets, such as urban environments captured over time and space. The lack of integration of spatiotemporal information limits the models' ability to capture dynamics over time and across spatial regions, especially in tasks requiring an understanding of both spatial and temporal dependencies. Therefore, there is a need for self-supervised learning approaches that effectively incorporate spatiotemporal information to enhance performance in such tasks.

\subsection{Spatiotemporal Contrastive Learning in Vision Tasks}
Spatiotemporal contrastive learning enhances traditional contrastive learning by integrating both spatial and temporal information, enabling models to capture underlying relationships in unlabeled data that vary over space and time.

Temporal contrastive learning excels in sequential data by differentiating between related and unrelated frames. For example, Contrastive Predictive Coding (CPC)~\citep{oord2019representation} applies temporal contrastive learning by using consecutive video frames as positive pairs and shuffled or temporally distant frames as negative pairs, helping models learn temporal coherence. 
SeCo~\citep{manas2021seasonal} and GeoSSL \citep{ayush2021geographyaware} use multi-season remote sensing images for self-supervised pre-training, enhancing model performance in remote sensing tasks.

Spatial contrastive learning improves a model's ability to represent spatial scenes from various angles, perspectives, and locations. Multi-view contrastive learning approach is typically applied within a single scene from multiple angles at one location~\citep{tian2020contrastive}.
Building on these concepts, geospatial contrastive learning contrasts data from different geographic locations or regions. By ensuring that data from similar spatial locations are closer in the feature space while data from different regions are more distant, models can more effectively capture spatial patterns and geographic features~\citep{deuser2023sample4geo,klemmer2024satclip,mai2023csp,guo2024spatialscene2vec}. This approach enhances the understanding of spatial relationships across wider geographic contexts.

\subsection{Street View Representation Learning for Downstream Tasks}
Street view imagery has been widely used in various urban downstream tasks, such as road defect detection~\citep{chacra2018municipal}, urban function recognition~\citep{huang2023comprehensive}, and socioeconomic prediction~\citep{fan2023urban}. However, existing research on street view representation often relies on supervised models trained on datasets like Places365~\citep{zhou2017places} or directly uses the pixel proportions of semantic segmentation results. These approaches fail to fully capture the rich semantic information embedded in street view imagery. Unlike natural images, street view imagery not only contains complex visual semantics but also encodes valuable spatiotemporal information in its metadata. Effectively representing this dual semantic nature -- both visual and spatiotemporal -- remains a significant challenge for improving its use in urban downstream tasks. Although a few studies have explored spatiotemporal self-supervised learning approaches to represent street view imagery~\citep{stalder2024selfsupervised}, these methods still have limitations. For instance, Urban2Vec~\citep{wang2020urban2vec} incorporates spatial information into self-supervised training by constructing positive sample pairs based on nearest neighbors, while KnowCL~\citep{liu2023knowledgeinfused} integrates knowledge graphs with contrastive learning to align locale and visual semantics, improving the accuracy of socioeconomic prediction using street view imagery. However, these approaches fail to explore the natural meanings of the spatiotemporal attributes of street view imagery and how to leverage these attributes to construct self-supervised methods suitable for various downstream tasks.

\section{Method}

The real world undergoes continuous changes across both temporal and spatial dimensions, yet these changes exhibit a certain level of continuity. In the temporal dimension, it is important to capture the invariant characteristics of a location as they evolve over time. Meanwhile, in the spatial dimension, the focus is on maintaining the consistency of the overall ambiance within a specific spatial range. These temporal and spatial invariances are crucial for enhancing performance in various downstream tasks. In this section, we introduce the proposed spatiotemporal contrastive learning framework in detail (Figure~\ref{fig:framework}).

Contrastive learning aims to learn feature representations from unlabeled data by contrasting positive and negative samples. The primary goal is to minimize the distance between positive samples and maximize the distance between negative samples within the feature space. Positive pairs are constructed by applying data augmentations to street view images. By optimizing the InfoNCE loss function, the model learns to reduce the distance between positive pairs in the feature space and increase the distance from negative samples, thus improving the feature representation learning. Given a query representation \(q\) and a set of positive and negative keys \((k^{+}, k^{-})\), the InfoNCE~\citep{oord2019representation} loss is defined as:

\begin{equation}
\mathcal{L}_{q} = - \log \frac{\exp \left(q \cdot k^{+} / \tau\right)}{\exp \left(q \cdot k^{+} / \tau\right)+\sum_{k^{-}} \exp \left(q \cdot k^{-} / \tau\right)}
\end{equation}

Here, \(q\) is the feature representation of the query, \(k^{+}\) is the feature representation of the positive sample, and \(k^{-}\) is the feature representation of negative samples. The temperature parameter \(\tau\) controls the scaling of similarities. The goal is to maximize the similarity between the query and the positive key \(q \cdot k^{+}\) while minimizing the similarity between the query and the negative keys \(q \cdot k^{-}\).
Building on this contrastive learning framework, we introduce temporal and spatial contrasts for constructing positive pairs from street view images.

\textbf{Temporal Contrastive Learning.}
Street view images captured at the same location but at different times differ from video frames because the intervals between shots are not fixed. Unlike remote sensing images, street view images taken at different times are not perfectly aligned in terms of geographic locations. Due to the typical spatial and angular shifts between images captured at different times, we impose restrictions on the conditions for positive temporal pairs: they must be taken within 5 meters of each other and have the same shooting angle. The historical street view image set for each location can be represented as \( T = [t_1, t_2, \dots, t_n] \), where \(t_i\) denotes the images captured at different times. Since the number of images varies for each location, the value of \(n\) differs accordingly. The aim of temporal contrast is to capture the invariant features of the same location over time. This means that even though the images are taken at different times, the model should learn to recognize the consistent characteristics of the scene.

To capture invariant features of the same location over time, we define the temporal contrastive loss. Given a positive sample pair \( (t_i, t_j) \) that meets temporal conditions (images taken within 5 meters and from the same angle), the temporal contrastive loss is:
\begin{equation}
    \mathcal{L}_{\text{t}} = - \log \frac{\exp \left(t_i \cdot t_j / \tau\right)}{\exp \left(t_i \cdot t_j / \tau\right) + \sum_{t_k^{-}} \exp \left(t_i \cdot t_k^{-} / \tau\right)}
\end{equation}
where \( t_i \) and \( t_j \) are feature representations of the positive temporal samples, \( t_k^{-} \) denotes negative samples from different locations or angles, and \( \tau \) is the temperature parameter for scaling. This formulation aims to maximize similarity between the same location’s images taken at different times while minimizing similarity to negatives.

\textbf{Spatial Contrastive Learning.}
Capturing the spatial consistency of an urban area is essential for accurately representing the urban physical environment. Spatial consistency refers to the ability to recognize that different locations within the same urban area still represent the same underlying physical characteristics. To achieve this, we treat all street view images captured within a specific urban area as representing the same environment, even if these images are taken from different angles or slightly different positions. This approach allows the model to account for variations in location while preserving the overall ambiance of the area. The set of street view images for a given urban area can be denoted as \(S = \{s_1, s_2, \dots, s_n\}\), where each \(s_i\) represents an image captured within the defined area. These images collectively provide a comprehensive spatial representation of the urban environment. We randomly select two samples \((s_i, s_j)\) from the set \(S\) and treat them as positive pairs. This encourages the model to learn that despite slight variations in shooting angle or position, the images are part of the same spatial context.

To capture spatial consistency within an urban area, we define the spatial contrastive loss. Given a set of street view images \( S = \{s_1, s_2, \dots, s_n\} \) from the same urban area, we randomly select two samples \( (s_i, s_j) \) as a positive pair and define the spatial contrastive loss as:
\begin{equation}
    \mathcal{L}_{\text{s}} = - \log \frac{\exp \left(s_i \cdot s_j / \tau\right)}{\exp \left(s_i \cdot s_j / \tau\right) + \sum_{s_k^{-}} \exp \left(s_i \cdot s_k^{-} / \tau\right)}
\end{equation}
where \( s_i \) and \( s_j \) are feature representations of the positive spatial samples, and \( s_k^{-} \) represents negative samples from different urban areas. This loss encourages the model to maximize similarity between images in the same urban area while minimizing similarity to negatives from other areas.
 By doing so, we enable the model to learn consistent and representative spatial features across the entire urban area.

\section{Experiments and Results}

To validate our hypothesis, we first pre-train the models using datasets specifically designed for self-supervised contrastive learning, spatial contrastive learning, and temporal contrastive learning, respectively. We then evaluate the models on three distinct downstream tasks that reflect the characteristics of these contrastive learning models: visual place recognition (VPR), socioeconomic indicator prediction, and safety perception. Additionally, we conduct interpretability analyses on the features learned by the different contrastive models to gain a deeper understanding of the information the models focus on and how this impacts performance on urban downstream tasks.

\subsection{Pre-training Datasets and Experiment Setup}
\label{sec:appendix_data}

To obtain street view imagery for both self-supervised model training and socioeconomic indicator prediction, we first sourced road network data for each city using the OSMnx library \citep{boeing2017osmnx} from OpenStreetMap. We then generated query points along these road networks at regular intervals of 15 meters. The Google Street View (GSV) Application Programming Interface (API) was subsequently utilized to retrieve and download street view images. 

Since the VPR and safety perception datasets include a wide range of street view images from different cities, while the socioeconomic prediction task focuses more on local city characteristics, we constructed two separate datasets — a global version and a local version — for testing on different downstream tasks.

For the global version, to capture a broad spectrum of urban environments, we trained our self-supervised models on data collected from ten diverse and representative global cities including Amsterdam, Barcelona, Boston–Cambridge–Medford–Newton (Boston), Buenos Aires, Dubai–Sharjah (Dubai), Johannesburg, Los Angeles, Melbourne, Seoul, and Singapore. These cities were carefully selected to encompass a variety of geographical locations, cultural backgrounds, and urban forms, ensuring the diversity and richness of our training dataset. We collected historical images of ten global cities from the Google Street View (GSV) API which resulted in a total of over 42 million street view images used for pre-training.

For the local version, we selected street view images from Los Angeles to construct different contrastive datasets tailored to the specific needs of the socioeconomic prediction task in that city. The construction methods of datasets are similar to the global version.

Based on the street view pre-training datasets, we constructed three distinct contrastive datasets corresponding to different contrastive learning models for both global and local versions: self-contrastive, temporal contrastive, and spatial contrastive datasets. To benchmark against the MoCov3 baseline trained on ImageNet, each dataset was standardized to consist of 1 million image pairs. This uniform dataset size facilitates a fair comparison among the models by ensuring that each receives an equal amount of training data.

\textbf{Self-contrastive Dataset}. For the self-contrastive dataset, we randomly selected 100,000 images from each of the 10 cities, resulting in a total of 1 million images. Positive pairs were generated during training by applying data augmentation techniques to these images, following the settings used in MoCo v3 \citep{chen2021empirical}. Additionally, for the local version, we constructed a self-contrastive dataset based solely on Los Angeles using the same method.

\textbf{Temporal Contrastive Dataset.} In constructing the temporal contrastive dataset, we randomly selected 100,000 street view sampling points from each of the 10 cities, totaling 1 million sampling points. At each sampling point, we retrieved images taken at different times but from the same shooting angle. Two images were randomly selected from the temporal sequence to form a positive pair, resulting in 1 million temporal positive pairs. Similarly to the self-contrastive dataset, we constructed an additional temporal contrastive dataset based solely on Los Angeles using the same method.

\textbf{Spatial Contrastive Dataset.} For the global spatial contrastive dataset, we defined a 100-meter buffer zone as a unified urban area. From each buffer zone, we randomly selected two images to form positive pairs. Out of all the spatial positive pairs generated, we then randomly selected 1 million pairs to create the spatial contrastive dataset. It is important to note that we did not impose any restrictions on the shooting angle for positive pairs, allowing the model to focus more on the overall ambiance of the urban environment rather than specific street layouts. Similarly, for the local version, since the socioeconomic dataset is based on block groups, we defined each block group as an urban area and constructed positive pairs based on the block group boundaries.

\textbf{Training.}
We use AdamW \citep{loshchilov2019decoupled} as the optimizer, a common choice for training ViT base~\citep{dosovitskiy2021an} models, with a weight decay of 1e-6. For each dataset, we use a mini-batch size of 1024 and an initial learning rate of 6e-6. The model is trained for 300 epochs, starting with a 40 epoch warmup\citep{goyal2018accurate}, followed by a cosine decay schedule for learning rate decay \citep{loshchilov2017sgdr}. Training the ViT Base model for 300 epochs on 4 Nvidia A800 GPUs takes approximately 71 hours.

\subsection{Visual Place Recognition}

\begin{figure*}[ht]
\begin{center}
    \includegraphics[width=\linewidth]{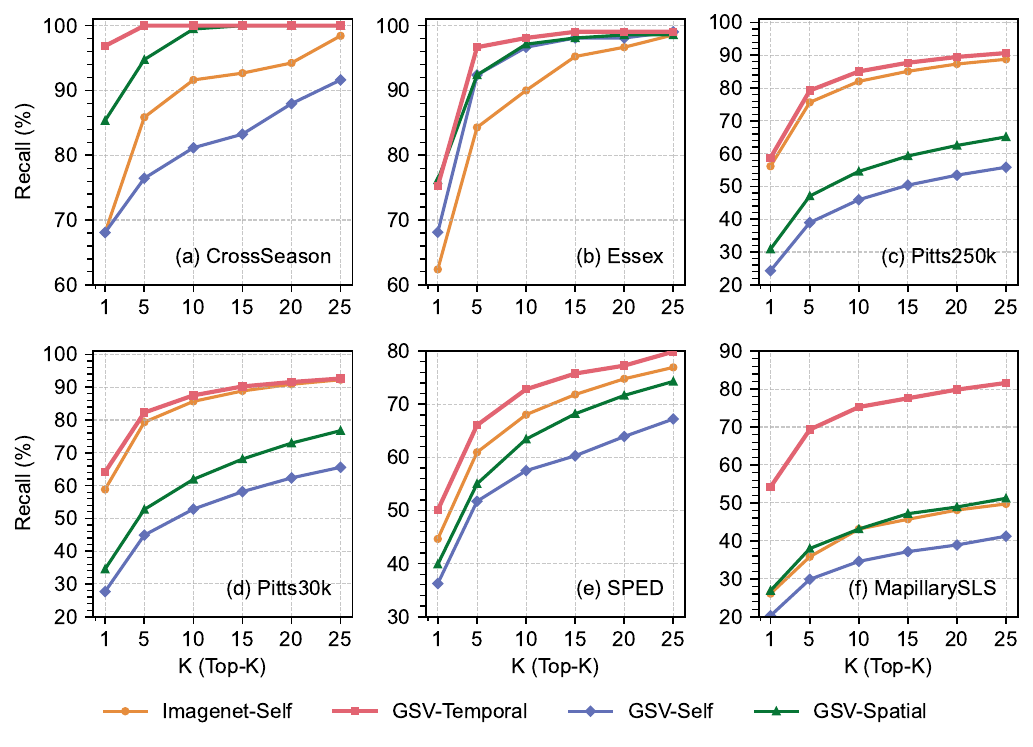}  
\end{center}
\caption{Performance comparison on different visual place recognition datasets (Recall@K in \%).}
\label{fig:vpr}
\end{figure*}

Visual place recognition (VPR) is a crucial urban task that aims to identify specific locations based on visual input. This task requires the removal of temporal disturbances to focus on stable information that does not change over time, demanding feature extraction that effectively distinguishes constant characteristics in the environment to improve recognition accuracy.

To evaluate the model's performance in VPR tasks, we used several benchmark datasets: CrossSeason~\citep{manslarsson2019crossseason}, Essex~\citep{zaffar2021memorable}, Pitts250k, Pitts30k~\citep{arandjelovic2018netvlada}, SPED~\citep{chen2018learning}, and MapillarySLS~\citep{warburg2020mapillary} datasets. Detailed information about these benchmark datasets are described in \ref{sec:appendix_vpr}. The models was tested by freezing the backbone of the pre-trained ViT and extracting the [CLS] token for VPR tasks. We assessed performance using the Recall@K metric, measuring the models' ability to correctly identify query image locations among the top-k most similar database images.

In Figure \ref{fig:vpr}, the GSV-Temporal model demonstrates exceptional performance on the CrossSeason dataset, achieving a recall value of 100\% across all K values. This indicates its robust capability in cross-season VPR tasks. In contrast, GSV-Self and ImageNet-Self exhibit significantly lower performance, suggesting their inability to effectively capture temporal invariant features.
On the Essex dataset, GSV-Temporal maintains a recall value exceeding 75\%, with values of 99.05\% for both K=20 and K=25. This highlights its not sensitivity to dynamic changes in the environment, outperforming other models in this context.
In the Pitts250k dataset, GSV-Temporal consistently outperforms GSV-Self and ImageNet-Self in recall values, the GSV-Temporal model also excels on the Pitts30k dataset, achieving a recall value of 90.23\% at K=15.  underscoring its suitability for complex urban environments in VPR tasks.
For the SPED dataset, GSV-Temporal displays superior recall values compared to other models, particularly with a notable performance at K=5. In the MapillarySLS dataset, GSV-Temporal showcases its outstanding performance again, with a recall value of 77.57\% at K=15.

In summary, the GSV-Temporal model consistently outperforms other models across multiple datasets, particularly in VPR tasks. Its not sensitivity to temporal and environmental changes positions it as a superior choice for this application, revealing significant potential for practical use.

\subsection{Socioeconomic Indicator Prediction}

\begin{table*}[htbp]
    \centering
    \caption{Model performance comparison on socioeconomic indicator prediction tasks based on LASSO across contrastive models.}
    \label{tab:detail_r2}
    \begin{tabular}{clcccc}
        \hline
        \textbf{Topic} & \textbf{Label} & \textbf{GSV-Self} & \textbf{GSV-Spatial} & \textbf{GSV-Temporal} & \textbf{ImageNet-Self} \\
        \hline
        \textbf{Crime} & Log(Violent Crime) & 0.4203 & \textbf{0.4287} & 0.4194 & 0.4146 \\
         & Log(Petty Crime) & 0.1810 & 0.1877 & \textbf{0.1892} & 0.1667 \\
          & Total & 0.3007 & \textbf{0.3082} & 0.3043 & 0.2906 \\
        \hline
        \textbf{Health} & \% Cancer Health & 0.6644 & \textbf{0.6969} & 0.6618 & 0.6053 \\
         & \% Diabetes & 0.6589 & \textbf{0.6942} & 0.6796 & 0.6172 \\
         & \% LPA & 0.8001 & \textbf{0.8337} & 0.8221 & 0.7671 \\
         & \% Mental Health & 0.7088 & \textbf{0.7510} & 0.7291 & 0.6753 \\
         & \% Obesity & 0.7628 & \textbf{0.7886} & 0.7797 & 0.7175 \\
         & \% Physical Health & 0.7120 & \textbf{0.7399} & 0.7314 & 0.6752 \\
          & Total & 0.7178 & \textbf{0.7507} & 0.7340 & 0.6763 \\
        \hline
        \textbf{Poverty} & Log(Income) & 0.6561 & \textbf{0.6816} & 0.6735 & 0.6096 \\
         & \% Poverty Line (100\%) & 0.1948 & \textbf{0.2227} & 0.1833 & 0.1718 \\
         & \% Poverty Line (200\%) & 0.6154 & 0.6377 & \textbf{0.6401} & 0.5893 \\
      & Total & 0.4888 & \textbf{0.5140} & 0.4990 & 0.4569 \\
        \hline
        \textbf{Transport} & \% Drive Alone & 0.3841 & \textbf{0.3991} & 0.3835 & 0.3582 \\
         & PMT & 0.6196 & \textbf{0.6447} & 0.6289 & 0.5379 \\
         & PTRP & 0.6024 & \textbf{0.6385} & 0.6087 & 0.5302 \\
         & VMT & 0.6647 & \textbf{0.6921} & 0.6874 & 0.6163 \\
         & VTRP & 0.6900 & \textbf{0.6994} & 0.6991 & 0.6436 \\
         & \%Public Transit & 0.5226 & \textbf{0.5700} & 0.5339 & 0.4726 \\
         & \%Walk & 0.2383 & \textbf{0.2925} & 0.2340 & 0.2080 \\
          & Total & 0.5317 & \textbf{0.5623} & 0.5394 & 0.4810 \\
        \hline
        \textbf{Overall Total} &  & 0.5609 & \textbf{0.5888} & 0.5714 & 0.5209 \\
        \hline
    \end{tabular}
\end{table*}

The socioeconomic indicator prediction task aims to use street view images to infer the socioeconomic status of urban areas. It emphasizes learning the overal ambiance of a region rather than specific geometric features, highlighting the need for feature extraction to focus on similarities between regions to better understand economic conditions and developmental dynamics.

In the downstream task of predicting socioeconomic indicators, we utilized the socioeconomic dataset published by~\citet{fan2023urban}, which contains 18 socioeconomic indicators across seven major cities in the United States (Table~\ref{table:indicators}). We take the socioeconomic indicator prediction of Los Angeles as an example. Detailed descriptions are provided in Section~\ref{sec:appendix_socio}. We first extracted street view features from the images using the pre-trained models of the local version. These features were then aggregated using the mean values at the block group level. The aggregated features were used as input features to predict socioeconomic indicators for each block group.

For prediction model training and evaluation, we split each city's dataset into a training set (70\%) and a testing set (30\%). We used LASSO as the regressor to evaluate the predictive performance of the image features extracted by the different pre-trained models. Additionally, we applied 5-fold cross-validation to ensure robust evaluation. This approach allows for a fair comparison of the different contrastive learning models in capturing visual features that are meaningfully correlated with socioeconomic indicators.

The results of socioeconomic indicator predictions are shown in Table \ref{tab:detail_r2}.
Overall, models pre-trained on street view images significantly outperform the model pre-trained on the ImageNet dataset. Specifically, across all 18 indicators, the ImageNet-pretrained model achieved an average \(R^2\) of 0.5209. 
In contrast, models on street view images achieved average \(R^2\) scores of 0.5609 for self-contrastive, 0.5714 for temporal contrastive, and 0.5888 for spatial contrastive models, respectively.
Furthermore, both temporal and spatial contrastive pre-training models capture more socioeconomic-related information compared to the self-contrastive approach, with spatial contrastive demonstrating the highest performance. This trend is consistent across most socioeconomic indicators, showing the strongest predictive performance for Health-related indicators and the least for Crime-related indicators.

These findings suggest that spatial contrastive pre-training effectively captures the overall ambiance of urban areas, enabling more precise predictions of regional socioeconomic information. Additionally, temporal contrastive pre-training filters out random factors and dynamic elements in the images, enhancing the reliability of socioeconomic predictions.

\subsection{Safety Perception}
The safety perception task involves using street view imagery to estimate how safe people perceive a given scene to be. To make accurate estimates, this task requires analyzing all relevant elements within the scene, as each can contribute to the overall perception of safety, particularly elements such as trees and vehicles \citep{zhang2018measuring}.

\begin{table*}[ht]
    \centering
    \caption{Evaluation metrics of different models on the safety perception classification task.}
    \begin{tabular}{lllll}
    \hline
    \multicolumn{1}{c}{\bf Model}  & \multicolumn{1}{c}{\bf Accuracy (\%)} & \multicolumn{1}{c}{\bf Recall (\%)} & \multicolumn{1}{c}{\bf F1 Score (\%)} & \multicolumn{1}{c}{\bf AUC Score (\%)} \\ \hline 
        ImageNet-Self   & 83.25            & 70.32          & 75.43             & 80.51                \\
        GSV-Temporal    & 84.91            & 65.16          & 75.94             & 80.72                \\
        GSV-Spatial     & 86.08            & 68.39          & 78.23             & 82.33                \\
        GSV-Self        & \textbf{88.68}   & \textbf{77.42} & \textbf{83.33}    & \textbf{86.29}      \\ \hline 
    \end{tabular}
    \label{tab:safety}
\end{table*}

We selected the PlacePlus 2.0 \citep{dubey2016deep} dataset for the downstream task of human environmental perception, filtering out over 1,144 images with safety perception scores below 3.5 and above 6.5, with 80\% of the data used for training and 20\% for testing. The model was trained using a linear binary classification approach for 20 epochs to effectively distinguish between low and high safety perception environments.

Table \ref{tab:safety} compares the performance of various models in classifying safety perception in urban environments. Notably, the GSV-Self model achieved the highest accuracy (88.68\%) and recall (77.42\%), demonstrating its effectiveness in identifying both safe and unsafe environments while minimizing false negatives. Its F1 score of 83.33\% indicates a strong balance between precision and recall, and the AUC score of 86.29\% further confirms its ability to distinguish between safety levels across thresholds. Overall, the GSV-Self model outperforms the others in all metrics, underscoring its potential for applications in urban safety perception tasks.

\subsection{Analysis of Differences in Spatio-Temporal Contrastive Features}

\begin{figure*}[ht]
\begin{center}
    \includegraphics[width=\linewidth]{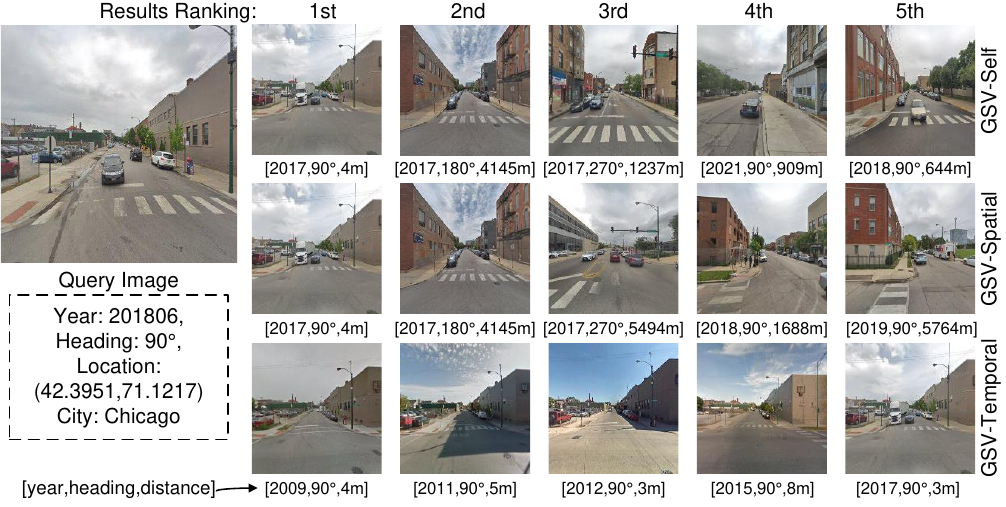}
\end{center}
\caption{
Comparison of retrieval results using GSV-Self, GSV-Spatial, and GSV-Temporal methods for a given query image (Year: 2018, Heading: $90^\circ$, Location: Chicago). Each row corresponds to 
the top-5 retrieved street view images based on different self-supervised pertained models, ranked by image feature similarity to the query image. 
The GSV-Temporal results are all within a 10-meter radius and have identical heading angles, but correspond to different time periods, demonstrating temporal invariance of the learned image representations. The GSV-Spatial results cover a larger geographic area with nearby timeframes, maintaining a consistent  overal ambiance.
}
\label{fig:query}
\end{figure*}
This section explores the differences in feature representation and retrieval tasks by comparing the Self, Spatial, and Temporal contrastive methods. We use street view images from Chicago, analyzing the performance of these three contrastive methods in pre-trained models. 
For each model pre-trained with a unique contrastive learning objective, we extract a 768-dimensional feature vector representing the characteristics of street view images. 
In our experiment, to comprehensively evaluate the retrieval performance of these methods, we randomly selected 500 street view images from different locations in Chicago as query images, ensuring that each image originated from a distinct spatial location. For each query image, we used the Nearest Neighbors method in feature space to retrieve the top five street view images with the closest Cosine distance. This process generated a total of 500 sets of query and retrieval pairs, with five results for each query. Specifically, we used the Euclidean distance as a similarity measure to rank and obtain the top five retrieval results.

Finally, we randomly selected one set of query and retrieval results for visualization. By comparing the year, heading, and feature distance of the retrieved results with the query image, we visually demonstrated the significant differences in retrieval characteristics among the three contrastive methods. Figure \ref{fig:query} shows the retrieval results for a given query street view image using GSV-Self, GSV-Spatial, and GSV-Temporal contrastive methods. On the left, the query image is displayed, including information about the year of capture (June 2018), heading (90°), geographic location (42.3951, 71.1217), and city (Chicago). The retrieval results are arranged in three rows, corresponding to the GSV-Self, GSV-Spatial, and GSV-Temporal methods (Figure 2). Each row shows the top five most similar retrieval results, ranked from left to right (1st to 5th). Below each retrieved image, the year of capture, heading, and actual distance from the query image (in meters) are indicated. The GSV-Self method retrieves the nearest street view images based on deep feature similarity. From the comparison, it can be seen that although the retrieved images are from different locations, they are very similar to the query image in feature space, indicating that GSV-Self emphasizes overall visual feature similarity without considering consistency in geographic location, heading, or time. The GSV-Spatial retrieval results cover a larger geographic area, allowing for greater spatial variation while aiming to maintain a similar overall ambiance and temporal proximity. It can be observed that most of the retrieved street view images are relatively dispersed in space, but the  overall ambiance and time are relatively close, reflecting spatial and environmental consistency. This allows GSV-Spatial to capture visually similar urban characteristics across different locations. The GSV-Temporal retrieval results maintain the same heading and are strictly limited to within a 10-meter radius, highlighting temporal diversity. While the position and heading are mostly unchanged, the retrieved images come from different years. This approach demonstrates sensitivity to temporal changes while keeping other factors consistent, thereby showcasing the variation of the same location across different years.

\subsection{What do GSV-Temporal and GSV-Spatial Contrastive Objectives Learn from GSV?}
Our experimental results reveal that different contrastive learning methods excel in different tasks: Temporal contrastive performs exceptionally well in VPR tasks, Spatial contrastive shows better results in macroeconomic prediction tasks, and Self contrastive achieves the best performance in safety perception tasks, confirming our hypothesis that street view images captured at the same location over time enable contrastive learning tasks to uncover the temporal-invariant characteristics of the built environment. Similarly, spatially proximate street view images from the same period facilitate learning tasks to capture the spatial-invariant neighborhood ambiance, such as socioeconomic overal ambiance. To further understand how different models allocate their attention to various aspects of the input, we visualized the attention maps in ViT and evaluated the spatial extent of attention using attention distance. This analysis reveals the distinct focus areas of each model, shedding light on their feature extraction preferences.

\begin{figure*}[ht]
\begin{center}
    \includegraphics[width=\linewidth]{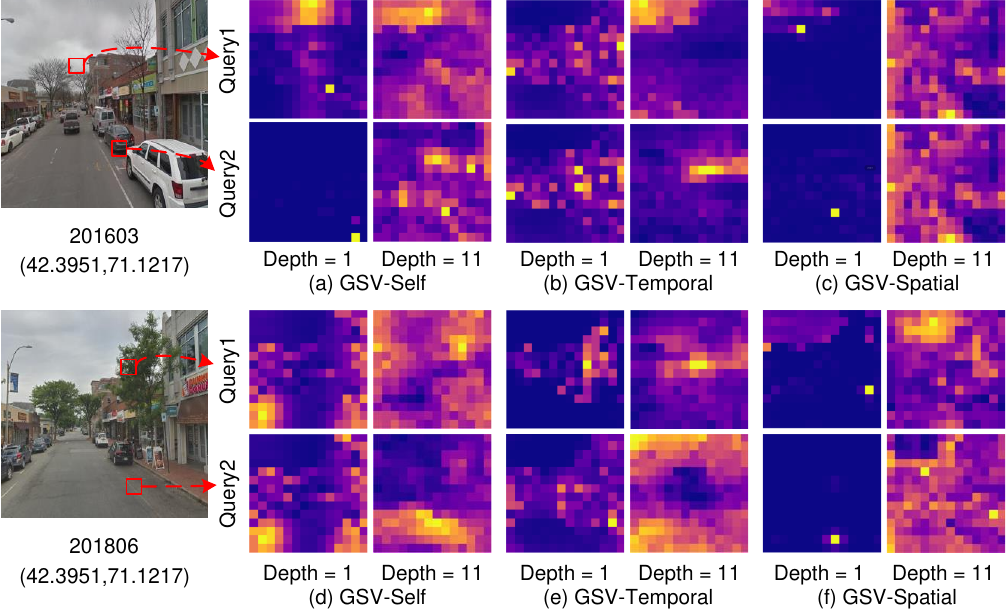}
\end{center}
\caption{Attention maps for two queries visualized across models and depths. Red boxes indicate regions of focus. GSV-Self (a, d) emphasizes objects like cars. GSV-Temporal (b, e) filters out dynamic objects, highlighting static elements. GSV-Spatial (c, f)  shows consistent focus across queries, capturing overall spatial structures.}
\label{fig:attention}
\end{figure*}

\textbf{GSV-Temporal learns temporal invariant characteristics, and GSV-spatial learns invariant neighborhood ambiance.}

This attention map visualization (Figure \ref{fig:attention}) shows how different contrastive learning strategies encode spatial and temporal invariants within urban street view images. The attention maps~\citep{chefer2021transformer} highlight how the models focus on distinct regions across various depths.
We selected two street view images of the same location taken at different times. The attention maps for two query tokens, marked in red on the images, were visualized across layers from the first to the last depth, and the detailed results are available in Section~\ref{sec:appendix_attention}.

In the first depth, GSV-Self and GSV-Temporal exhibit a broader distribution of attention, while GSV-Spatial focuses more on localized regions. This suggests that GSV-Self and GSV-Temporal prioritize capturing global information in the early stages, whereas GSV-Spatial tends to emphasize detailed information initially.
However, in the last depth, GSV-Self (Figure~\ref{fig:attention}a, d) attends to global information across the image but tends to focus more on regions near the query token. In contrast, the GSV-Temporal model (Figure~\ref{fig:attention}b, e) shows that query 1 (placed in the sky) primarily attends to the sky, filtering out dynamic elements. Query 2, placed on a car (a dynamic object), shows no attention to the car, reinforcing the model's ability to learn temporal-invariant characteristics by ignoring dynamic elements.
In the GSV-Spatial model (Figure \ref{fig:attention}c, f), both query 1 and query 2 show similar attention patterns across the images. The model focuses on the overall structure without emphasizing dynamic objects like cars, indicating that spatial contrastive learning effectively captures spatial-invariant environmental characteristics. This supports the hypothesis that spatial contrast learning emphasizes the broader environment rather than individual objects.



\begin{figure}[ht]
\begin{center}
    \includegraphics[width=\linewidth]{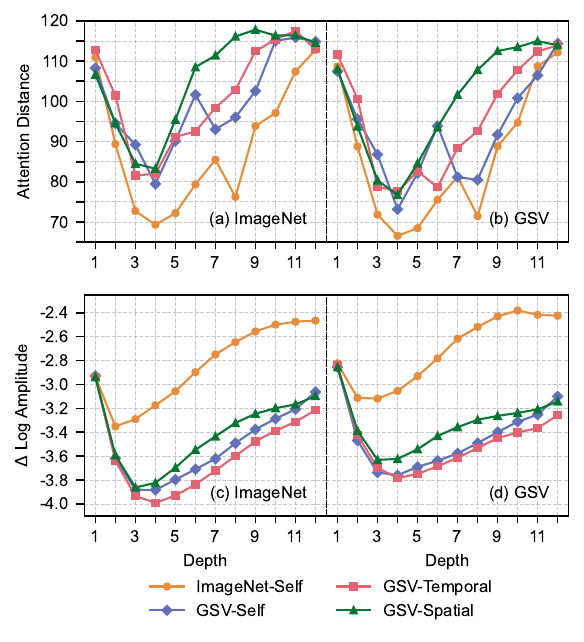}
\end{center}
\caption{Visualization of attention distance and $\Delta$ Log Amplitude across depths for ImageNet and GSV models. Depth refers to the network layers in the ViT model, from shallow (Depth 1) to deep layers (Depth 12). (a) and (b) display the attention distance, which represents the average spatial range of the attention mechanism in each layer—a larger value indicates that the model attends to more globally distributed features, while smaller values suggest a focus on local patterns. (c) and (d) present the $\Delta$ Log Amplitude, where higher values (closer to 0) reflect stronger retention of high-frequency information (e.g., edges, textures), and lower values (more negative) indicate a focus on low-frequency components, representing global structures or smooth transitions. }
\label{fig:at_amp}
\end{figure}

We evaluate the spatial extent of self-attention using attention distance~\citep{dosovitskiy2021an}, which measures the mean distance between query tokens and key tokens, weighted by their respective self-attention scores. This metric helps assess how different contrastive strategies focus on various aspects of the scene. Figure \ref{fig:at_amp}(a) and \ref{fig:at_amp}(b) show the attention distances computed for sampled street view images and ImageNet images. Depth corresponds to the network layers in the ViT model, ranging from shallow (Depth 1) to deep layers (Depth 12). Larger attention distances indicate that the model captures more globally distributed features, while smaller distances suggest a focus on local patterns.
Specifically, GSV-Spatial exhibits the largest attention distance, indicating a tendency to focus on a broader spatial context rather than concentrating on individual objects. In contrast, the attention distances of GSV-Temporal and GSV-Self decrease sequentially, suggesting a gradual narrowing of focus to capture more specific details within the scenes. Notably, ImageNet-Self demonstrates the smallest attention distance, reflecting its pre-training on a dataset primarily consisting of object-centric images, which leads to a greater emphasis on individual objects over the overall spatial arrangement.

\textbf{GSV-Temporal highlights low-frequencies, and GSV-spatial exploits high-frequencies.}

The low-frequency amplitude of an image represents its large-scale structure and smooth transitions, primarily encompassing the background, gradient regions, and general contours. It reflects the overall form of the image and broad variations in brightness. Low-frequency components are typically key elements in global structure modeling and scene consistency understanding, which is why their amplitude is generally larger. In contrast, the high-frequency amplitude of an image represents finer details, textures, and edges, and is primarily associated with regions of rapid changes in the image, such as boundaries and local contrast variations. Although high-frequency amplitudes are relatively smaller, they are crucial for capturing the sharpness and clarity of the image and often contain noise signals.
In this study, we hypothesize that, compared to GSV-Spatial, GSV-Temporal is more inclined to focus on low-frequency information. This is because temporal-invariant characteristics in street view images rely more on global consistency and stable structures, while high-frequency information are more susceptible to noise interference in dynamic scenes. To test this hypothesis, we compute the amplitude differences in the Fourier-transformed frequency spectrum of intermediate features across various layers of the ViT backbone, reporting the relative amplitudes of high and low frequencies~\citep{park2023what}. Specifically, Figures \ref{fig:at_amp}(c) and \ref{fig:at_amp}(d) present the relative amplitude results for ImageNet and GSV images under different contrast strategies.

The results indicate that model pre-trained on ImageNet focus more on high-frequency information, while models pre-trained on GSV emphasize low-frequency information. This difference may stem from the fact that ImageNet images typically center around object categories (e.g., animals, plants, and etc.) that require detailed edge and texture detection, thus highlighting high-frequency information. In contrast, street view images feature large-scale street layouts and global structural variations, where the models need to capture more low-frequency information to understand the overall spatial relationships within the scene. Furthermore, we observe that GSV-Temporal exhibits the most pronounced sensitivity to low-frequency information. This suggests that the temporal-invariant characteristics prioritize the consistency of static elements, such as street layouts, while being less sensitive to texture variations caused by factors like lighting or seasonality. GSV-Self, similar to GSV-Temporal, also focuses more on low-frequency information, but due to the need to capture dynamic elements such as pedestrian and vehicular flow, it exhibits a slightly higher relative amplitude compared to GSV-Temporal. On the other hand, GSV-Spatial shows a stronger focus on high-frequency information. This can be attributed to its lesser sensitivity to the overall street layout, as it is more concerned with capturing consistency in the surrounding environment, which is often conveyed through high-frequency details such as window styles, building facades, and material textures.








\section{Conclusion}
In this work, we propose a self-supervised urban visual representation learning framework based on street view images, capable of selectively extracting and encoding dynamic and static information and their ambiance in urban environments according to the requirements of different downstream tasks. By leveraging the unique spatiotemporal attributes of street view imagery, we have developed three contrastive learning strategies: temporal invariance representation, spatial invariance representation, and global information representation. Experimental results demonstrate that these strategies can effectively learn task-specific features suitable for their respective downstream applications, significantly enhancing performance in urban environment understanding tasks. Furthermore, we conducted an in-depth analysis of the reasons behind the performance of different contrastive methods, further emphasizing the importance of targeted learning strategies. This study systematically explores representation learning strategies based on street view images, provides a valuable benchmark for the application of visual data in urban science, and enhances their practical applicability.

\section{Acknowledgements}
This work was supported by the High-performance Computing Platform of Peking University. We also acknowledge the financial support from the National Natural Science Foundation of China (Grant No. 42371468).

 \bibliographystyle{elsarticle-harv} 
 \bibliography{reference.bib}



\appendix
\renewcommand{\thefigure}{A\arabic{figure}} 
\renewcommand{\thetable}{A\arabic{table}}   
\setcounter{figure}{0} 
\setcounter{table}{0}  

\section{Data}


\subsection{Socioeconomic indicator prediction dataset}
\label{sec:appendix_socio}

In our downstream task, we used socioeconomic indicators provided by \citet{fan2023urban}, which include data from seven major metropolitan areas in the United States. We take Los Angeles as an example. The socioeconomic indicators cover various topics relevant to urban studies and are detailed in Table \ref{table:indicators}.

\begin{table*}[ht]
\caption{Socioeconomic Indicators}
\label{table:indicators}
\begin{center}
\begin{tabular}{lll}
\hline
\multicolumn{1}{c}{\bf Topic} & \multicolumn{1}{c}{\bf Indicator} & \multicolumn{1}{c}{\bf Label} \\
\hline 
Crime & Violent crime occurrence per spatial unit & Log(Violent Crime) \\

& Violent theft-related crime occurrence per spatial unit & Log(Petty Crime) \\

\hline

Health & Model-based estimate for crude prevalence of  & \% Cancer Health \\
& cancer (excluding skin cancer) among adults aged $\geq 18$ years & \\

& Model-based estimate for crude prevalence of  & \% Diabetes \\
& diagnosed diabetes among adults aged $\geq 18$ years & \\

& Model-based estimate for crude prevalence of  & \% LPA \\ 
& no leisure-time physical activity among adults aged $\geq 18$ years  & \\

& Model-based estimate for crude prevalence of  & \% Mental Health \\
& mental health not good for $\geq 14$ days among adults aged $\geq 18$ years  & \\

& Model-based estimate for crude prevalence of  & \% Obesity \\ 
& obesity among adults aged $\geq 18$ years  & \\

& Model-based estimate for crude prevalence of  & \% Physical Health \\ 
& physical health not good for $\geq 14$ days among adults aged $\geq 18$ years  & \\
\hline
Poverty & Median Household Income & Log(Income) \\

 & \% Individuals with poverty status determined:  & \% Poverty Line (100\%) \\
 & below 100\% poverty line & \\
 
 & \% Individuals with poverty status determined: & \% Poverty Line (200\%) \\
 & below 200\% poverty line & \\
 
\hline
Transport & \% Population (\textgreater 16) commute by driving alone  & \% Drive Alone \\

& Estimated personal miles traveled on a working weekday   & PMT \\

& Estimated personal trips traveled on a working weekday   & PTRP \\

& Estimated vehicle miles traveled on a working weekday   & VMT \\ 

& Estimated vehicle trips traveled on a working weekday   & VTRP \\ 

& \% Population (\textgreater 16) commute by public transit   & \%Public Transit \\

&  \% Population (\textgreater 16) commute by walking and biking  & \%Walk \\
\hline
\end{tabular}
\end{center}
\end{table*}

\subsection{Visual Place Recognition dataset}
\label{sec:appendix_vpr}

\paragraph{ESSEX.}
The ESSEX dataset provides a diverse set of urban and suburban scenes with varying viewpoints and lighting conditions. It challenges the model's robustness in recognizing places despite changes in perspective and environmental factors~\citep{zaffar2021memorable}.
\paragraph{CrossSeason:}
 This dataset contains images captured across different seasons, aiming to study the impact of seasonal variations on image features. It is primarily used to train and evaluate models for visual recognition under varying seasonal conditions~\citep{manslarsson2019crossseason}.
\paragraph{Pittsburgh:}
 This is a large-scale dataset featuring street view images from Essex in the UK and Pittsburgh in the USA. It is designed to support visual localization and geographic scene recognition tasks, providing rich environmental diversity suitable for various urban analysis studies~\citep{arandjelovic2018netvlada}.
\paragraph{SPED:}
This dataset focuses on the temporal changes in street view imagery, containing images of the same location captured at different time points. It aims to study the dynamic features of urban environments, suitable for temporal analysis and scene change detection~\citep{chen2018learning}.
\paragraph{MapillarySLS:}
 This dataset includes street view images from around the globe, designed to support tasks in autonomous driving and visual understanding. Generated by users, it covers a variety of environments and conditions, providing rich geographical and scene information~\citep{warburg2020mapillary}.

\section{Interpretation}
\label{sec:appendix_attention}
\begin{figure*}[ht]
\begin{center}
    \includegraphics[width=\linewidth]{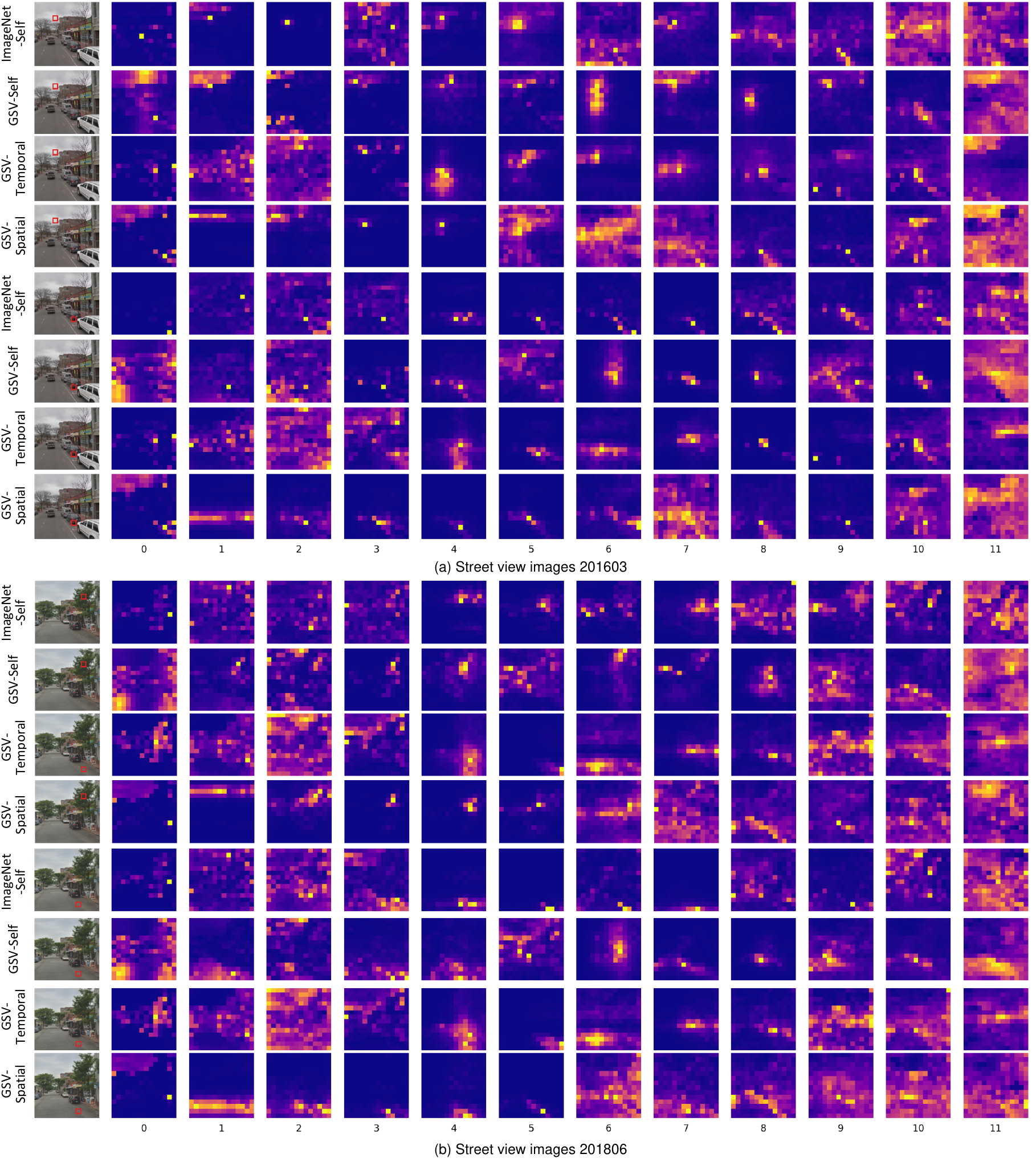}  
\end{center}
\caption{Attention maps for two queries visualized across models and depths.}
\label{fig:attention_all}

\end{figure*}

We visualized the attention maps across all depths for each contrastive learning strategy. The rows correspond to different strategies—ImageNet-Self, GSV-Self, GSV-Temporal, and GSV-Spatial—while the columns represent different depths (0 to 11). The original street view inputs are displayed on the left. Each attention map highlights the regions of the image that the model focuses on, demonstrating how attention shifts across depths for self, temporal and spatial features (Figure \ref{fig:attention_all}).

\end{document}